\def\year{2022}\relax
\pdfoutput=1
\documentclass[letterpaper]{article} 
\usepackage{aaai22}  
\usepackage{times}  
\usepackage{helvet}  
\usepackage{courier}  
\usepackage[hyphens]{url}  
\usepackage{graphicx} 
\usepackage{natbib}  
\usepackage{caption} 
\DeclareCaptionStyle{ruled}{labelfont=normalfont,labelsep=colon,strut=off} 
\usepackage{algorithmic}

\title{Inferring Probabilistic Reward Machines from Non-Markovian Reward Signals for Reinforcement Learning}

\author{
    Taylor Dohmen\textsuperscript{\rm 1}\thanks{These authors contributed equally},
   Noah Topper\textsuperscript{\rm 2}\footnotemark[1],
   George Atia\textsuperscript{\rm 2},
   Andre Beckus\textsuperscript{\rm 3},\\
   Ashutosh Trivedi\textsuperscript{\rm 1},
   Alvaro Velasquez\textsuperscript{\rm 3} \\
}
\affiliations{
    \textsuperscript{\rm 1} University of Colorado Boulder \quad
    \textsuperscript{\rm 2} University of Central Florida \quad
    \textsuperscript{\rm 3} Air Force Research Laboratory
}

\usepackage{xcolor}
\usepackage{amsmath}
\usepackage{amssymb}
\usepackage{mathtools}

\usepackage{amsthm}
\newtheorem{lemma}{Lemma}
\newtheorem{theorem}{Theorem}

\newtheorem{example}{Example}
\newtheorem{definition}{Definition}

\usepackage[linesnumbered, ruled, noend]{algorithm2e}
\usepackage{hyperref}
\hypersetup{hidelinks}
\usepackage{cleveref}

\newcommand{\bb}[1]{\mathbb{#1}}

\newcommand{\set}[1]{\left\{ #1 \right\}}

\newcommand{\paren}[1]{\left( #1 \right)}

\newcommand{\AP}{\mathsf{AP}}

\newcommand{\sem}[1]{[\![#1]\!]}

\renewcommand{\vec}[1]{\mathbf{#1}}

\newcommand{\pr}[1]{\Pr \left[ #1 \right]}

\newcommand{\tn}[1]{\textnormal{#1}}

\newcommand{\rank}[1]{\mathsf{rank}\left( #1 \right)}

\newcommand{\rep}[1]{\mathsf{rep}\left( #1 \right)}

\newcommand{\compat}[2]{#1 \sim #2}

\newcommand{\compatRow}[2]{#1 \stackrel{\scriptscriptstyle E}{\sim} #2}
\newcommand{\notCompatRow}[2]{#1 \stackrel{\scriptscriptstyle E}{\nsim} #2}

\DeclareMathOperator*{\argmax}{\arg\max}

\newcommand{\diff}[3]{\mathsf{diff}_{#1}\left( #2, #3 \right)}

\usepackage{tikz}
\usetikzlibrary{
  patterns,
  calc,
  angles,
  quotes,
  automata,
  positioning,
  decorations.markings,
  arrows,
  intersections,
  shapes,
  arrows,
  backgrounds
}

\colorlet{darkgreen}{green!40!black}
\colorlet{darkblue}{blue!60!black}
\colorlet{darkred}{red!50!black}
\colorlet{safecellcolor}{yellow!5}
\colorlet{goodcellcolor}{green!10}
\colorlet{badcellcolor}{blue!10}

\tikzset{
  >=stealth,
  box state/.style={draw,rectangle,minimum size=8mm},
  prob state/.style={draw,very thick,shape=circle,darkblue,minimum size=3mm,inner sep=0mm},
  node distance=2cm,on grid,auto, initial text=,
  every loop/.style={shorten >=0pt},
  accepting/.style={double distance=1.2pt, outer sep = 0.6pt+\pgflinewidth},
  accepting dot/.style={above=-2.5pt,circle,fill,darkgreen,inner sep=2pt,radius=1pt},
  loop above/.append style={every loop/.append style={out=120, in=60, looseness=6}},
  loop below/.append style={every loop/.append style={out=300, in=240, looseness=6}},
  loop left/.append style={every loop/.append style={out=210, in=150, looseness=6}},
  loop right/.append style={every loop/.append style={out=30, in=330, looseness=6}},
  accepting arc/.style={dashed},
  marked/.style={
    dashed,
    opacity=0.3
  },
  marked on/.style={alt=#1{marked}{}},
}

\newcommand{\mdp}{\textnormal{MDP}}
\newcommand{\tmdp}{\textnormal{TMDP}}
\newcommand{\prm}{\textnormal{PRM}}

\begin{document}

\maketitle

\begin{abstract}
The success of reinforcement learning in typical settings is predicated on Markovian assumptions on the reward signal by which an agent learns optimal policies.
In recent years, the use of reward machines has relaxed this assumption by enabling a structured representation of non-Markovian rewards.
In particular, such representations can be used to augment the state space of the underlying decision process, thereby facilitating non-Markovian reinforcement learning.
However, these reward machines cannot capture the semantics of stochastic reward signals.
In this paper, we make progress on this front by introducing probabilistic reward machines (PRMs) as a representation of non-Markovian stochastic rewards.
We present an algorithm to learn PRMs from the underlying decision process and prove results around its correctness and convergence.
\end{abstract}

\section{Introduction}
\label{sec:intro}
Traditionally, reinforcement learning (RL) \cite{SuttonBarto98} techniques have relied on strong Markovian assumptions of the underlying decision process.
In particular, the reward signal by which an agent learns through positive or negative reinforcement is generally defined as a function of the current state of the environment and the action of an agent.
The history of states observed by the agent are not considered in such settings.
However, many problem domains require taking into account this history.
Examples include learning in settings with sparse rewards \cite{NeiderGaglioneGavranTopcuWuXu21}, rewards defined as regular expressions and formal logics \cite{CamachoIcarteKlassenValenzanoMcIlraith19}, and decision-making under partial observability \cite{IcarteWaldieKlassenValenzanoCastroMcIlraith19}.
In such settings, the non-Markovian reward signal may not be known, but can be learned from traces of behavior.
Recently, the problem of learning a structured representation from the reward signal has received significant attention.
These representations often take the form of finite automata called reward machines \cite{IcarteKlassenValenzanoMcIlraith18,IcarteKlassenValenzanoMcIlraith22}.
Reward machines can augment the state space of decision processes with non-Markovian rewards and thereby facilitate the use of classical techniques for Markov Decision Process (\mdp{}) on these more general types of environments \cite{GaonBrafman20, XuWuOjhaNeiderTopcu21}.
They also serve as a memory mechanism for reasoning over partially observable environments \cite{IcarteWaldieKlassenValenzanoCastroMcIlraith19}, are useful for reward shaping to mitigate sparse reward signals \cite{CamachoIcarteKlassenValenzanoMcIlraith19,VelasquezMelcer20,VelasquezBisseyBarakBeckusAlkhouriMelcerAtia21}, and provide explanations of RL systems \cite{XuWuOjhaNeiderTopcu21}.

While much progress has been made on learning and leveraging reward machines for decision processes with non-Markovian rewards \cite{XuGavranAhmadMajumdarNeiderTopcu20,XuWuOjhaNeiderTopcu21,AbadiBrafman20,GaonBrafman20,NeiderGaglioneGavranTopcuWuXu21,RensRaskinReynouardMarra21}, the more general setting where rewards exhibit both non-Markovian and stochastic dynamics has not been addressed.
In this paper, we make progress on this front by introducing \emph{probabilistic reward machines} (\prm{}s).

We present an algorithm to learn \prm{} representations of non-Markovian stochastic reward signals by integrating a probabilistic variant of the $L^\star$ algorithm \cite{Angluin87} for automaton inference with RL.
Our algorithm uses a product construction to produce a standard \mdp{} from the environmental decision process and the learned \prm{}s in order to perform RL, and we provide an accompanying proof that these products correctly simulate the original decision processes with non-Markovian stochastic rewards.
To further motivate \prm{}s, we show that an exponential blowup in size is required to emulate the system under learning by embedding probabilities in the environment and using deterministic reward machines.
We establish theoretical guarantees on convergence of the algorithm in the limit.



\subsection*{Related Work}
The classical problem of grammatical inference seeks to learn a formal language, or some representation thereof, from a finite number of samples \cite{Horning69, DeLaHiguera10}.
Approaches for such learning problems are generally categorized as either active or passive.
Passive inference seeks to mine the underlying specification from a static dataset of observed traces.
Active methods differ in that the system under learning can be queried to guide the learning process.
The $L^\star$ algorithm for learning regular languages \cite{Angluin87} is the quintessential example of active inference, and assumes the existence of a minimally adequate teacher capable of answering membership and equivalence queries.
This method has been broadly adopted and generalized to learn interface automata \cite{AartsVaandrager10}, Mealy machines \cite{Niese03}, automaton representations of recurrent neural networks \cite{WeissGoldbergYahav18,WeissGoldbergYahav19}, and \mdp{}s \cite{TapplerAichernigBacciEichlsederLarsen19}.

Grammatical inference has become a popular problem in the context of RL and learning reward signals.
While most literature assumes a reward machine is given \cite{IcarteKlassenValenzanoMcIlraith18,HahnPerezScheweSomenziTrivediWojtczak19}, the problem of learning the machine from observations has only been considered recently.
The work in \cite{XuGavranAhmadMajumdarNeiderTopcu20,NeiderGaglioneGavranTopcuWuXu21} explores a solving approach based on satisfiability.
In \cite{IcarteWaldieKlassenValenzanoCastroMcIlraith19}, the problem of learning a reward machine is viewed through the lens of discrete optimization.
Methods that merge Q-learning and $L^\star$ for learning reward machines are considered in \cite{GaonBrafman20,XuWuOjhaNeiderTopcu21}.

\section{Probabilistic Reward Machines}
\label{sec:prm}
We proceed with our treatment on \prm{}s with some preliminary definitions pertaining to the underlying decision process from which samples are obtained to estimate the \prm{} representation of the objective.

A \emph{Transition-Markov Decision Process} (\tmdp{}) maintains the classical assumption that state transitions are Markovian while generalizing the standard \mdp{} model by allowing history-dependent stochastic rewards.

\begin{definition}[\tmdp{}]
A \emph{transition-Markov decision process} $M$ is a tuple $(X, x_I, A, P, \AP, L, R)$ where $X$ is a finite set of states, $x_I \in X$ is an initial state, $A$ is a finite set of actions, $P : (X \times A \times X) \to [0,1]$ is a probabilistic transition function, $\AP$ is a finite set of atomic propositions, $L : (X \times A \times X) \to 2^\AP$ is a function labelling each transition with a subset of $\AP$, and $R : \paren{2^\AP}^* \to \mathsf{Dist}(\bb{R})$ is a non-Markovian stochastic reward function, where $\mathsf{Dist}(\bb{R})$ denotes the set of possible probability distributions over $\bb{R}$.
\end{definition}

A run of a \tmdp{} is a string $x_0 a_1 x_1 \cdots a_n x_n \in X \cdot (A \cdot X)^*$ such that $0 < \prod^n_{k=1} P(x_{k-1}, a_k, x_k)$.
For a run $x_0 a_1 x_1 \cdots a_n x_n$, its corresponding label sequence is $\ell_1 \ell_2 \cdots \ell_n$, where $\ell_k = L(x_{k-1}, a_k, x_k)$, and the corresponding reward sequence is $r_1 r_2 \cdots r_n$ where $r_k \sim R(\ell_1 \cdots \ell_k)$.
A trajectory is a run for which $x_0 = x_I$.
A policy is a function $\pi : \paren{X \cdot (A \cdot X)^* \times A} \to [0,1]$ specifying a distribution over the set of actions following a run.

Given a \tmdp{} $M = (X, x_I, A, P, \AP, L, R)$, we seek to construct a \prm{} that encodes the underlying reward $R$.

\begin{definition}[\prm{}]
A \emph{probabilistic reward machine} $H$ is a tuple 
$(\AP, \Gamma, Y, y_I, \tau, \varrho)$ where $\AP$ is a set of atomic propositions and $2^\AP$ is the input alphabet, $\Gamma \subset \bb{R}$ is a finite set of rewards, $Y$ is a finite set of states, $y_I \in Y$ is a distinguished initial state, $\tau : \paren{Y \times 2^\AP \times Y} \to [0,1]$ is a probabilistic transition function, and $\varrho : \paren{Y \times 2^\AP \times Y} \to \Gamma$ is a function mapping each transition to a reward from $\Gamma$.
\end{definition}

\begin{figure}
    \centering
    \resizebox{\linewidth}{!}{%
    \begin{tikzpicture}[node distance = 3cm]
        \node[state,initial] (0) {$y_0$};
        \node[state] (2) [below of = 0] {$y_2$};
        \node[state] (1) [left of = 2] {$y_1$};
        \node[state] (3) [right of = 2] {$y_3$};
        \node[state] (4) [below of = 2] {$y_4$};
        
        \path[->] (0) edge [loop right] node {$\neg c \wedge \neg * \mid 1$} node [left] {0} (0);
        \path[->] (0) edge node [sloped,above] {$c \wedge \neg * \mid 0.9$} node [sloped,below] {0} (1);
        \path[->] (0) edge node [sloped,above] {$c \wedge \neg * \mid 0.1$} node [sloped,below] {0} (3);
        \path[->] (0) edge node [left] {$* \mid 1$} node [right] {0} (2);
        
        \path[->] (2) edge [loop below] node {$\tn{true} \mid 1$} node [above] {0} (2);
        
        \path[->] (1) edge [loop left] node [sloped,above,pos=0.75] {$\neg o \wedge \neg * \mid 1$} node [right] {0} (1);
        \path[->] (1) edge node [above] {$* \mid 1$} node [below] {0} (2);
        \path[->] (1) edge node [below,sloped] {$o \mid 1$} node [above,sloped] {1} (4);
        
        \path[->] (3) edge [loop right] node [sloped,below,pos=0.75] {$\neg o \wedge \neg * \mid 1$} node [left] {0} (3);
        \path[->] (3) edge node [above] {$* \mid 1$} node [below] {0} (2);
        \path[->] (3) edge node [below,sloped] {$o \mid 1$} node [above,sloped] {0} (4);
        
        \path[->] (4) edge [loop right] node {$\tn{true} \mid 1$} node [left] {0} (4);
    \end{tikzpicture}
    }
    \caption{A \prm{} defined over the office gridworld.}
    \label{fig:coffeeworldexample}
\end{figure}
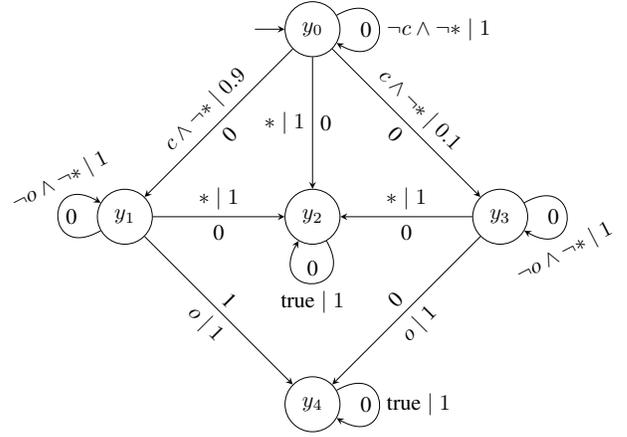

A run of $H$ is a sequence $y_0 (\ell_1, \gamma_1) y_1 \cdots (\ell_n, \gamma_n) y_n$ where $\gamma_k = \varrho(y_{k-1}, \ell_k, y_k)$ and $\tau(y_{k-1}, \ell_k, y_k) > 0$, for all $k$.
Let $\vec{H} : \paren{2^\AP \cdot \Gamma} \to [0,1]^{|Y| \times |Y|}$ map each pair  $\ell, \gamma \in 2^\AP \times \Gamma$ to the transition matrix $\vec{H}(\ell \gamma)$ such that
\begin{equation*}
	\vec{H}(\ell \gamma)[i,j] = \begin{cases} 
		\tau(y_i, \ell, y_j) & \tn{if } \varrho(y_i, \ell, y_j) = \gamma, \\
		0 & \tn{otherwise.}
	\end{cases}
\end{equation*}
As usual, we extend the domain from $2^\AP \cdot \Gamma$ to $\paren{2^\AP \cdot \Gamma}^*$ by setting $\vec{H}(\ell \gamma w) = \vec{H}(\ell \gamma) \vec{H}(w)$ and $\vec{H}(\varepsilon) = \vec{I}$, where $\vec{I}$ is the identity matrix and $\varepsilon$ the empty string.
A distribution $\bb{P}_H : \paren{(2^\AP)^* \times \Gamma^*} \to [0,1]$ over pairs of inputs and reward sequences is induced:
\begin{equation*}
    \bb{P}_H(\ell_1 \cdots \ell_n, r_1 \cdots r_n) = \vec{y}_I \vec{H}(\ell_1 r_1 \cdots \ell_n r_n) \vec{1},
\end{equation*}
where $\vec{y}_I$ is an initial distribution over $Y$ (typically with an entry of 1 at the index for state $y_I$ and zeros everywhere else) and $\vec{1}$ is the vector with 1 at each entry.
The \prm{} $H$ is said to encode the reward function $R$ of a \tmdp{} if, for any label sequence $\ell = \ell_1 \cdots \ell_n$ and reward sequence $r = r_1 \cdots r_n$, the following equality holds:
\begin{equation*}
	\bb{P}_H(\ell, r) = \prod^n_{k=1} R(\ell_1 \cdots \ell_k)(r_k).
\end{equation*}

\begin{example}
Consider the \textit{office gridworld} environment found in \cite{IcarteKlassenValenzanoMcIlraith18}.
The original task, which takes the form of a deterministic reward machine, requires that the agent acquire coffee at location \textit{c} and delivers the coffee to office \textit{o}, at which point the agent receives a reward of 1.
If the agent steps on an $*$, they fail the task and observe a reward of 0.
A variant of this task, in which the reward signal is probabilistic, may be obtained by introducing a 10\% chance that the coffee machine malfunctions, producing weak coffee which is rejected upon delivery and receives a reward of 0.
A graphical representation of a \prm{} for this task is shown in \Cref{fig:coffeeworldexample}.
An edge $y \xrightarrow[\scriptscriptstyle r]{\scriptscriptstyle \ell | p} y'$ indicates that $y$ transitions to $y'$ on label $\ell \in 2^\AP$ with probability $p=\tau(y, \ell, y')$ and receives reward $r=\varrho(y, \ell, y')$.
\end{example}

The following property formalizes the notion of a \prm{} in which every state and label uniquely determines the reward.

\begin{definition}[Reward-Determinism]
\label{def:reward-determinism}
    A \prm{} is \emph{reward-deterministic} if, for any $y, y_1, y_2 \in Y$ and $\ell \in 2^\AP$ such that $\tau(y, \ell, y_1) > 0$ and $\tau(y, \ell, y_2) > 0$, it holds that either $y_1 = y_2$ or $\varrho(y, \ell, y_1) \neq \varrho(y, \ell, y_2)$.
\end{definition}

\section{An Excursus on $L^\star$}
\label{sec:lStar}
In the seminal work of \cite{Angluin87}, an algorithm called $L^\star$ is proposed by which a deterministic finite automaton (DFA) for a given regular language $\mathcal{L}$ can be learned through the use of a minimally adequate teacher that can answer membership and equivalence queries.
When the learner executes a membership query, it presents a word $w \in (2^\AP)^*$ to the teacher and the teacher outputs whether $w \in \mathcal{L}$.
If the learner executes an equivalence query, it presents a hypothesis DFA to the teacher who must then answer whether this automaton encodes the language to be learned.
If not, the teacher generates a counterexample in the form of a word on which the two languages differ.
The $L^\star$ algorithm keeps track of state words $S \subseteq (2^\AP)^*$, which are closed under prefix operations (e.g. if $ab \in S$, then $a \in S$) and test words $E \subseteq (2^\AP)^*$, which are closed under suffix operations (e.g. if $ab \in E$, then $b \in E$).
Initially, we have $S = E = \{\varepsilon\}$, where $\varepsilon$ is the empty string.
As the algorithm proceeds, there are two critical properties that must be tracked and revolve around the notion of $E$-equivalence.

\begin{definition}[$E$-Equivalence]
Given $w, w' \in (2^\AP)^*$ and set $E \subseteq (2^\AP)^*$, the words $w$ and $w'$ are $E$-equivalent with respect to the language $\mathcal{L}$, denoted $w \equiv_E w'$, if $w  e \in \mathcal{L} \iff w'  e \in \mathcal{L}$ holds for every $e \in E$.
\end{definition}


\begin{definition}[Consistency]
Given $(S, E)$, the consistency property holds if and only if for all $s, s' \in S$, the following implication holds: if $s \equiv_E s'$, then $s \ell \equiv_E s' \ell$ for all $\ell \in \Sigma$.
\end{definition}

\begin{definition}[Closedness]
Given $(S, E)$, the closedness property holds if and only if for all $s \in S$ and $\ell \in 2^\AP$, there exists some $s' \in S$ such that $s  \ell \equiv_E s'$.
\end{definition}

For a closed and consistent $(S, E)$, a corresponding automaton can be derived by taking each $E$-equivalence class of $S$ to be a state, with the empty string $\varepsilon$ as the starting state.
The transition function is defined using the closedness property. That is, whenever we have $s  \ell \equiv_E s'$ per the closedness property, then we also have the transition from $s$ to $s'$ upon observing the label $\ell$ in the DFA.
Furthermore, it follows from the consistency property that $s'$ is unique.
The accepting states are those in the language of the teacher.

The $L^\star$ algorithm ensures that $(S, E)$ remains closed and consistent until the target language is learned. This is done by constructing a DFA from $(S, E)$.
If a counterexample is produced, it is used to modify $(S, E)$ by adding the counterexample and all its prefixes to $S$. Membership queries are then used to determine the values of the new entries.




We now consider again the RL setting where we wish to learn the \prm{} representation of the objective as given by the reward signal, and mention some challenges that arise.
Indeed, note that the seemingly innocuous membership queries must now be answered indirectly through the \tmdp{}.
This introduces two challenges to the adoption of the $L^\star$ algorithm within RL.
First, there is no obvious way to perform membership queries.
The construction of a meaningful $w$ is not obvious on the part of the learner and answering this query is difficult on the part of the teacher.
This is because we can only reason about these words indirectly through the \tmdp{}, whereas traditional approaches reason directly over the target language.

The second challenge lies in performing equivalence queries, since the teacher cannot answer whether the currently learned \prm{} encodes the reward of the underlying \tmdp{}.
However, it is worth noting that counterexamples are obtained naturally through the RL process.
These counterexamples are words that either have not been seen before or whose corresponding reward sequence differs from other encounters of the same word.
The latter case arises due to the stochastic nature of the reward signal.
Thus, the answers to equivalence queries are obtained implicitly through RL.

\section{Explicating Probabilistic Machines from Non-Markovian Reward Signals}
\label{sec:learning}


We now describe an algorithm that learns a \prm{} representation of the stochastic reward function of a TMDP.

\begin{definition}[Sampling Observation Table]
    For an underlying \prm{} $(\AP, \Gamma, Y, y_I, \tau, \varrho)$, an observation table is a tuple $(S, E, T)$, where $S \subset (2^\AP \cdot \Gamma)^*$ is a prefix-closed set of \emph{samples}, $E \subset (2^\AP \cdot \Gamma)^* \cdot 2^\AP$ is a suffix-closed set of \emph{experiments}, and $T : S \cdot \left((2^\AP \cdot \Gamma) \cup \set{\varepsilon}\right) \cdot E \to (\Gamma \to \bb{N})$ 
    is a function mapping label-reward words to reward frequencies.
\end{definition}

Defining appropriate analogs of closedness and consistency for this type of observation table relies on a notion of statistical difference, based on the Hoeffding bound.

\begin{definition}
\label{def:difference}
    We say that the sequences $s, s' \in S \cdot 2^\AP$ are \emph{statistically different}, written $\diff{f}{s}{s'}$, with respect to a function $f : (2^\AP \cdot \Gamma)^* \cdot 2^\AP \to (\Gamma \to \bb{N})$ if either
    \begin{itemize}
    	\item $0 < N = \sum_{g \in \Gamma} f(s)(g)$ and $0 < N' = \sum_{g \in \Gamma} f(s')(g)$,
    	\item or there exists $\gamma \in \Gamma$ such that we have the inequality $\sqrt{\frac{1}{2} \ln \frac{2}{C}} \left(\sqrt{\frac{1}{N}} + \sqrt{\frac{1}{N'}}\right) < \left| \frac{f(s)(\gamma)}{N} - \frac{f(s')(\gamma)}{N'} \right|$,	where $C$ is a data-dependent confidence level.
			\footnote{A good choice~\cite{TapplerAichernigBacciEichlsederLarsen19} for $C$ is $\frac{1}{M^3}$ where $M$ is the number of samples taken thus far.}
    \end{itemize}
\end{definition}

\begin{algorithm*}[t]
	\DontPrintSemicolon
	\SetFuncSty{scshape}
	\SetKwFunction{MQ}{MembershipQuery}
	\SetKwFunction{EQ}{EquivalenceQuery}
	\KwOut{Observation table $O$ and/or \prm{} $H$.}
Initialize observation table $O = (S, E, T)$\;
\Repeat{there exists no counterexample $\chi$}{
	\label{alg1:line2}
		\If{exists counterexample $\chi = \ell_1 \gamma_1 \cdots \ell_n \gamma_n$}{
			\label{alg1:line3}
			$Prefixes \gets \set{\ell_1 \gamma_1 \cdots \ell_k \gamma_k : 1 \leq k \leq n}$\;
			$Queries \gets Prefixes \cdot \left( (2^\AP \cdot \Gamma) \cup \set{\varepsilon} \right) \cdot E$\;
				
			$S \gets S \cup Prefixes$\;
		}
	\Repeat{$O$ is closed and consistent}{
		\label{alg1:line7}
		\uIf{$O$ is not consistent}{
			\label{alg1:line8}
				Choose $s, s' \in S$, $\ell \in 2^\AP$, $\gamma \in \Gamma$, $e \in E$ such that $\compatRow{s}{s'}$ and $\diff{T}{s \ell \gamma e}{s' \ell \gamma e}$.\;
				$E \gets E \cup \set{\ell \gamma e}$\;
				$Queries \gets Queries \cup \left( S \cdot \left( (2^\AP \cdot \Gamma) \cup \set{\varepsilon} \right) \cdot \set{\ell \gamma e} \right)$\;
				\label{alg1:line11}
			}
			\ElseIf{$O$ is not closed}{
				\label{alg1:line12}
				Choose $s \in S$, $\ell \in 2^\AP$, $\gamma \in \Gamma$ such that $T(s \ell)(\gamma) > 0$ and $\notCompatRow{s \ell \gamma}{r}$ for all $r \in \rep{S}$.\;
				$S \gets S \cup \set{s \ell \gamma}$\;
				$Queries \gets Queries \cup \left( \set{s \ell \gamma} \cdot \left( (2^\AP \cdot \Gamma) \cup \set{\varepsilon} \right) \cdot E \right)$\;
				\label{alg1:line15}
			}
		\For{$\zeta \in Queries$}{
			\label{alg1:line16}
			 $T \gets$ \MQ{$\zeta$, $T$}\;
			 \label{alg1:line17}
		}
		\label{alg1:line18}
		$Queries \gets \emptyset$\;
	}
	\label{alg1:line19}
	$T, \chi \gets$ \EQ{$O$}\;
	\label{alg1:line20}
}
\label{alg1:line21}
\caption{$L^\star$-based Active Inference for \prm{} representation of Stochastic Non-Markovian Reward}
\label{algo:main}
\end{algorithm*}

Two cells $s e$ and $s' e'$ of an observation table $(S, E, T)$ are compatible, denoted $\compat{s e}{s' e'}$, if $\diff{T}{s e}{s' e'}$ is false.
We say that two rows $s, s'$ are compatible, denoted $\compatRow{s}{s'}$ if $\compat{s e}{s' e}$ holds for all $e \in E$.
Based on these relations, $S$ can be partitioned into \emph{compatibility classes} which are groups of statistically similar samples.
Unlike in traditional $L^\star$, where each sample translates to a state in the derived automaton, each state of the derived \prm{} will correspond to a representative from a compatibility class. Each class is defined via compatibility to its representative.
Note that the compatibility relation is not an equivalence relation; a given sample may be compatible with several representatives and two elements of a compatibility class may be incompatible.
To resolve these ambiguities, define the function $\mathsf{rank} : S \to \bb{N}$ as $\rank{s} = \sum_{\ell \in 2^\AP} \sum_{\gamma \in \Gamma} T(s \ell)(\gamma)$, to quantify the amount of information about each sample contained in the table.
The unique representative of $s$ is then defined as the compatible sample of maximal rank and is characterized by the function $\mathsf{rep} : S \to S$, given as $\rep{s} = \argmax_{s' \in S} \{\rank{s'} : \compatRow{s}{s'}\}$.
Denote by $\rep{S}$ the set $\set{\rep{s} : s \in S}$ of all representatives.

\begin{definition}[Sampling Closedness]
	An observation table is closed if, for all $s \ell \gamma \in S \cdot 2^\AP \cdot \Gamma$ with $T(s \ell)(\gamma) > 0$, there exists $r \in \rep{S}$ such that $\compatRow{s \ell \gamma}{r}$ holds.
\end{definition}

\begin{definition}[Sampling Consistency]	
	An observation table is consistent, if, for all compatible pairs in the set $\{(s, s') \in S \times S : \compatRow{s}{s'}\}$ and all $\ell, \gamma \in 2^\AP \times \Gamma$, either $T(s \ell)(\gamma) = 0$ or $T(s' \ell)(\gamma) = 0$, or $\compatRow{s \ell \gamma}{s' \ell \gamma}$.
\end{definition}

With the above definitions established, we are now in position to describe the high-level flow of the learning procedure, which is specified explicitly in \cref{algo:main}.
While the details of the sub-procedures are significantly different from those involved in traditional $L^\star$, the overall structure of the process is quite similar.
The algorithm starts with an empty observation table and enters the loop at line \ref{alg1:line2}.
Since the table is empty to begin with and there does not exist a counterexample, the conditional at line \ref{alg1:line3} and the inner loop are bypassed and an equivalence query is executed at line \ref{alg1:line20}.
Deferring the details for now, the equivalence query builds a hypothesis \prm{} from the observation table and challenges the teacher to find a counterexample showing a discrepancy between the hypothesis and the true reward function.
If no counterexample is found, the halting condition at line \ref{alg1:line21} is met and the algorithm terminates.

If a counterexample is found, it is returned and stored in the variable $\chi$ and we return to the beginning of the loop.
In this case, the conditional block at line \ref{alg1:line3} is entered and the counterexample $\chi$ and all of its prefixes are added to the set of samples $S$ of the observation table.
Additionally, we would like to gain more information about the counterexample, so we save it and its extensions in the variable $Queries$.

Next, the loop at line \ref{alg1:line7} is entered, in which the table is checked for consistency.
If the table is inconsistent (lines \ref{alg1:line8}-\ref{alg1:line11}), a witness $\ell \gamma e$ is found, added to the set $E$ of experiments, and the set of all samples extended by this witness is added to the set $Queries$.
If the table is consistent, we proceed (lines \ref{alg1:line12}-\ref{alg1:line15}) by checking if it is closed.
If it is not closed, we similarly find a witness, this time $s \ell \gamma$, add it to the sample set $S$ and add its extensions by the set of all experiments to the set $Queries$.
For each string in $Queries$ a membership query is executed to gather more data into the table (lines \ref{alg1:line16}-\ref{alg1:line17}).
The variable $Queries$ is then emptied (line \ref{alg1:line18}), and this sub-process (lines \ref{alg1:line7}-\ref{alg1:line19}) is iterated until the observation table is both closed and consistent.
At this point, a hypothesis \prm{} can be constructed again, and the entire process repeats until the equivalence query fails to produce a counterexample to the current hypothesis.

    The reward-determinism property (c.f. \cref{def:reward-determinism}) is inspired by the label-determinism property used for learning \mdp{}s in \cite{TapplerAichernigBacciEichlsederLarsen19}.
    \Cref{algo:main} always learns a reward-deterministic \prm{}, even if the true reward is specified as a non-reward-deterministic \prm{}.

\subsection{Equivalence Queries}
In order to construct a hypothesis \prm{}, we need a statistically significant amount of data about system trajectories to be stored in the observation table.
We assume that a threshold parameter $N \in \bb{N}$ specifying the minimum number of samples necessary to estimate a transition probability is provided.
We use $\bot$ to denote a sink state to which transitions are diverted when the data is insufficient for estimating transition probabilities otherwise. 

\begin{definition}[Hypothesis Reward Machine]
\label{def:hypothesis}
    Given a closed and consistent observation table $(S, E, T)$, we construct a \prm{} $H = (\AP, \Gamma, Y, y_I, \tau, \varrho)$ as follows:
    \begin{itemize}
        \item $Y = (\rep{S} \times \Gamma) \cup \set{y_I, \bot}$ with $y_I = (\varepsilon, 0)$,
        \item $\tau(y, \ell, y') = \begin{cases}
            0 & \tn{if } y = \bot \tn{ and } y' \neq \bot, \\
            1 & \tn{if } y = y' = \bot, \\
            1 & \tn{if } \sum_{\Gamma} T(y \ell)(g) < N, \\
            &\quad \tn{and } y' = \bot \\
            & \tn{if } \sum_{\Gamma} T(y \ell)(g) \geq N, \\
            \frac{T(y \ell)(\gamma)}{\sum_{\Gamma} T(y \ell)(g)} &\quad y' = (\rep{r \ell \gamma}, \gamma), \\
            &\quad \tn{and } y = (r, \eta),
        \end{cases}$
        \item $\varrho(y, \ell, y') = \begin{cases}
            0 & \tn{if } y = \bot \tn{ or } y' = \bot, \\
            \gamma & \tn{if } y' = (\rep{r \ell \gamma}, \gamma) \tn{ and } y = (r, \eta).
        \end{cases}$
    \end{itemize}
\end{definition}

\begin{algorithm}[tb]
    \DontPrintSemicolon
    \SetFuncSty{scshape}
    \SetKwFunction{EQ}{EquivalenceQuery}
    \SetKwFunction{Query}{Query}
    \SetKwProg{Fn}{Function}{:}{}
    \KwData{$N_{stop}$}
    \Fn{\EQ{$O$}}{
			Make hypothesis $H$ from table $O$, initialize $Q_H$.\;
			\label{alg2:line2}
    	$flag \gets$ \texttt{false}\;
    	\For{$1 \leq i \leq N_{stop}$}{
    		$\lambda, Q_H, flag \gets$ \Query{$Q_H, H, \tn{equiv}$}\;
				\label{alg2:line5}
    		\For{prefix $\ell_1 r_1 \cdots \ell_k r_k$ of $\lambda$}{
					\label{alg2:line6}
    		    $\lambda_k \gets \ell_1 r_1 \cdots \ell_k$\;
			    $T(\lambda_k)(r_k) \gets T(\lambda_k)(r_k) + 1$\;
					\label{alg2:line8}
    		}
    		\lIf{$flag$}{\label{alg2:line9}\KwRet $T$, $\lambda$}
    	}
    	\KwRet $T$, \texttt{None}\;
			\label{alg2:line10}
	}
    \caption{Equivalence Query}
    \label{algo:equiv}
\end{algorithm}

Equivalence queries are one of two interfaces between the learner and the teacher.
In our setting, the teacher does not have complete knowledge of the system under learning, and so the equivalence query amounts to repeatedly executing the system and the hypothesis in tandem, trying to find inconsistencies between the two.
\Cref{algo:equiv} specifies the procedure, which takes a closed and consistent observation table as its single parameter.
It begins on line \ref{alg2:line2} by constructing a hypothesis \prm{} according to \cref{def:hypothesis} and initializing the corresponding Q-table $Q_H$.
The variable $flag$ is initially set to \texttt{false} and will be set to true if the teacher finds a counterexample, i.e. a label-reward string impossible under the hypothesis \prm{}, including those leading to the failure state $\bot$.
Additionally, each call to the teacher (line \ref{alg2:line5}) returns a label-reward string $\lambda$ and an updated Q-table $Q_H$.
Regardless of whether $\lambda$ is a counterexample, its data is added to the observation table (lines \ref{alg2:line6}-\ref{alg2:line8}).
Next (lines \ref{alg2:line9}-\ref{alg2:line10}), we check if $flag$ is set to \texttt{true}, indicating $\lambda$ is a counterexample, and if so, the updated table is returned along with $\lambda$, ending the loop early.
Otherwise, the loop repeats, and halts after $N_{stop}$ iterations without encountering a counterexample.
In this case, the updated table is returned and the ``counterexample'' returned is \texttt{None}.

\subsection{Membership Queries}
Membership queries are the second interface between learner and teacher, and are designed to gather data on strings that are witnesses to the observation table's openness or inconsistency.
Following~\cite{XuWuOjhaNeiderTopcu21}, we perform a membership query for each such witness string using a pseudo-``reward machine'', defined below.

\begin{definition}[Membership Query Machine]
\label{def:member_PRM}
    Given a query string $\zeta = \ell_1 \gamma_1 \cdots \gamma_{n-1} \ell_n$, the corresponding membership query machine is $H_\zeta = (\AP, \Gamma, Y, y_I, \tau, \varrho)$ where
    \begin{itemize}
        \item $Y = \set{y_0, y_1, \ldots, y_n}$ with $y_I = y_0$,
        \item $\tau(y_k, \ell, y') = \begin{cases}
            1 & \tn{if } \ell = \ell_{k+1} \tn{ and } y' = y_{k+1} \\
            &\quad \tn{or if } \ell \neq \ell_{k+1} \tn{ and } y' = y_k, \\
            0 & \tn{otherwise,}\end{cases}$
        \item $\varrho(y_k, \ell, y') = \begin{cases}
            1 & \tn{if } \ell = \ell_{k+1} \tn{ and } y' = y_{k+1}, \\
            0 & \tn{otherwise.}
        \end{cases}$
    \end{itemize}
\end{definition}

This reward machine incentivizes exploration along the query trace. \Cref{algo:member} gives the membership query specification, which takes as parameters a query string $\zeta$ and the map $T$ from the observation table.
First (line \ref{alg3:line2}), a membership \prm{} $H_\zeta$ is constructed for $\zeta$, according to \cref{def:member_PRM} and a corresponding Q-table $Q_\zeta$ is initialized .
Next, the learner passes the membership \prm{} and the Q-table to the teacher (line \ref{alg3:line4}) and is given back a label-reward string $\lambda$ and the updated Q-table.
As in equivalence queries, the data contained in $\lambda$ is added to the observation table (lines \ref{alg3:line5}-\ref{alg3:line7}), and the process repeats $N_{query}$ times, where $N_{query}$ is a contextual parameter, known a-priori.
Finally, the modified observation table is returned (line \ref{alg3:line8}).

\begin{algorithm}[t]
    \DontPrintSemicolon
    \SetFuncSty{scshape}
    \SetKwFunction{MQ}{MembershipQuery}
    \SetKwFunction{Query}{Query}
    \SetKwProg{Fn}{Function}{:}{}
    \KwData{$N_{query}$}
    \Fn{\MQ{$\zeta$, $T$}}{
			Construct query machine $H_\zeta$ and initialize $Q_\zeta$.\;
			\label{alg3:line2}
		\For{$1 \leq i \leq N_{query}$}{
			$\lambda, Q_\zeta \gets$ \Query{$Q_\zeta, H_\zeta, \tn{member}$}\;
			\label{alg3:line4}
			\For{prefix $\ell_1 r_1 \cdots \ell_k r_k$ of $\lambda$}{
				\label{alg3:line5}
			    $\lambda_k \gets \ell_1 r_1 \cdots \ell_k$\;
			    $T(\lambda_k)(r_k) \gets T(\lambda_k)(r_k) + 1$\;
			}
			\label{alg3:line7}
		}
		\KwRet $T$\;
		\label{alg3:line8}
	}
    \caption{Membership Query}
    \label{algo:member}
\end{algorithm}

\subsection{RL-Based Teacher}

\begin{algorithm*}[t]
	\DontPrintSemicolon
	\SetFuncSty{scshape}
	\SetKwFunction{EGA}{EpsilonGreedyAction}
	\SetKwFunction{Step}{Step}
	\SetKwFunction{Cex}{CounterExample}
	\SetKwFunction{Query}{Query}
	\SetKwProg{Fn}{Function}{:}{}
\KwData{$N_{ep}$, $M = (X, x_I, A, P, \AP, L, R)$, $\epsilon$, $\beta$, $\alpha$}
\Fn{\Query{$Q, H, type$}}{
	$x, y, \lambda, flag \gets x_I, y_I, \varepsilon, \texttt{false}$\;
	\For{$1 \leq i \leq N_{ep}$}{
		$a \gets$ \EGA{$Q, x, y, \epsilon$}\;
		\label{alg4:nextaction}
		$x', r \gets$ \Step{$M$, $x$, $a$}\;
		$\lambda \gets \lambda \cdot L(x, a, x') \cdot r$\;
		$y', \gamma \gets$ \Step{$H$, $y$, $L(x, a, x')$}\;
		\If{$type$ = \tn{member}}{
			$Q(y, x, a) \gets (1 - \alpha) Q(y, x, a) + \alpha \paren{\gamma + \beta \max_{a' \in A} Q(y', x', a')}$\;
			\label{alg4:Q1}
		}
		\ElseIf{$type$ = \tn{equiv}}{
			$Q(y, x, a) \gets (1 - \alpha) Q(y, x, a) + \alpha (r + \beta \max_{a' \in A} Q(y', x', a'))$\;
			\label{alg4:Q2}
			\If{$\lambda$ is a counterexample}{
				$flag \gets$ \texttt{true}
			}
		}
		$x, y \gets x', y'$\;
	}
	\If{$type$ = \tn{member}}{
		\KwRet $\lambda, Q$\;
	}
	\ElseIf{$type$ = \tn{equiv}}{
		\KwRet $\lambda, Q, flag$\;
	}
}
\caption{Generic RL-driven Query}
\label{algo:qRL}
\end{algorithm*}

The function \textsc{Query} used in both equivalence and membership queries belongs to the teacher and encapsulates the RL portion of the algorithm.
It leverages the fact that the product of a TMDP and a \prm{} results in a standard \mdp{}.

\begin{definition}
\label{def:product}
Let $M = (X, x_I, A, P, \AP, L, R)$ be a TMDP and let $H = (\AP, \Gamma, Y, y_I, \tau, \varrho)$ be a \prm{}.
The product $M \times H$ is an \mdp{} $(X', x_I', A, P', \AP, L', R')$ with a Markovian reward function $R': 2^\AP \to \Gamma$ such that:
\begin{itemize}
	\item $X' = X \times Y$ with $x_I' = (x_I, y_I)$,
	\item $P' ((x, y), a, (x', y')) = P(x, a, x') \cdot \tau(y, L(x, a, x'), y')$,
	\item $L'((x, y), a, (x', y')) = L(x, a, x')$, and
    \item $R'(L'((x, y), a, (x', y'))) = \varrho(y, L(x, a, x'), y')$.
\end{itemize}
\end{definition}

Any convergent RL algorithm working on the product of the TMDP and membership reward machine, will converge to a policy generating the desired membership query with optimal probability.
The RL algorithm optimizes over the product of the TMDP and the membership reward machine, and continues to observe the reward signals from the \prm{}. In doing so, the policy that maximizes these rewards is guided towards answering the underlying membership query by finding a trajectory of states in the decision process whose induced trace of labels answers the membership query.
Similarly, any convergent RL algorithm working on the product of the TMDP and hypothesis reward machine, will converge to an optimal policy with respect to the rewards consistent with the hypothesis reward machine. If the hypothesis correctly encodes the true reward function, we get an optimal policy for the environment.

The RL-driven query is given in \cref{algo:qRL}, which takes a Q-table, a reward machine, and $type$ indicating whether the query is for membership or equivalence.
Furthermore, the procedure assumes access to contextual parameters $\epsilon$ for choosing the next action (line \ref{alg4:nextaction}), $\alpha$ and $\beta$ as the learning rate and discount factor, respectively, used in the Q-table updates (lines \ref{alg4:Q1} and \ref{alg4:Q2}), and $N_{ep}$ as the episode length of each RL execution.
Note that the teacher interacts with the environmental TMDP and the given \prm{} individually, but by doing so in tandem, it simulates the execution of their product.
It treats both $M$ and $H$ as black-boxes in the sense that it may only observe the current state and outputs of each after executing them with particular inputs using the \textsc{Step} function.
It is also assumed that the environment $M$ resets to its initial state after the \textsc{Query} function returns.

\section{Correctness, Convergence, \& Compactness}
\label{sec:theorems}
In this section we establish theoretical properties related to \prm{}s and the learning procedure presented above.
In particular we show that the product construction of \cref{def:product} preserves \tmdp{} semantics, prove that \cref{algo:main} converges in the limit, and establish that \tmdp{}s with \prm{} encoded reward functions are more compact than \tmdp{}s with deterministic reward machines when representing semantically equivalent systems.

\begin{definition}[\tmdp{} Semantics]
\label{def:TMDP_semantics}
    The semantics of a \tmdp{} $M = (X, x_I, A, P, \AP, L, R)$ is a function $\sem{M} : A^* \to \mathsf{Dist}\paren{(2^{\AP})^* \times \bb{R}}$ such that
    \begin{equation*}
        \sem{M}(w)(\ell, r) = \begin{cases}
            1 &\tn{if } w = \varepsilon \tn{ and } \ell = \varepsilon, \tn{ and } r = 0 \\
            \bot &\tn{if } w, \ell \tn{ is not observable} \\
            p \cdot q &\tn{if } w, \ell \tn{ is observable with prob. } p   \\
            &\quad \tn{and } q = R(\ell)(r)
        \end{cases}
    \end{equation*}
    Define the language $\mathcal{L}(M)$ of the \tmdp{} $M$ as the set $\set{(w, \ell) \in A^* \times (2^\AP)^* : \exists r \in \bb{R}.\ \sem{M}(w)(\ell, r) \neq \bot}$.
\end{definition}

\begin{definition}[\mdp{} Semantics]
\label{def:mdp_semantics}
    The semantics of an \mdp{} $M = (X, x_I, A, P, \AP, L, R)$ is a function $\sem{M} : \paren{A^* \times (2^{\AP})^*} \to \mathsf{Dist}(\bb{R})$ such that
    \begin{equation*}
        \sem{M}(w, \ell)(r) = \begin{cases}
            1 &\tn{if } w = \varepsilon \tn{ and } \ell = \varepsilon \tn{ and } r = 0 \\
            \bot &\tn{if } w,\ell \tn{ is not observable} \\
            p &\tn{if } w,\ell \tn{ is observable with prob. } p \\
            &\quad \tn{and } \ell = \ell'l \tn{ and } r = R(l)
        \end{cases}
    \end{equation*}
    Define the language $\mathcal{L}(M)$ of the \mdp{} $M$ as the set $\set{(w, \ell) \in A^* \times (2^\AP)^* : \exists r \in \bb{R}.\ \sem{M}(w,\ell)(\gamma) \neq \bot}$.
\end{definition}

Two \mdp{}s $M, M'$ are equivalent iff, for all $w \in A^*$ and $\ell \in (2^\AP)^*$, it holds that $\sem{M}(w,\ell) = \sem{M'}(w,\ell)$.

\begin{theorem}
    If $M$ is a \tmdp{} with a reward encoded by a \prm{} $H$, then, for all $w \in A^*$, $\ell \in (2^\AP)^*$, and $\gamma \in \Gamma$:
    \begin{equation*}
        \sem{M}(w)(\ell, \gamma) = \sem{M {\times} H}(w, \ell)(\gamma).
    \end{equation*}
\end{theorem}
\begin{proof}
    Suppose that $M = (X, x_I, A, P, \AP, L, R)$ is a \tmdp{}, $H = (\AP, \Gamma, Y, y_I, \tau, \varrho)$ is a \prm{} encoding $R$, and $M {\times} H = (X', x_I', A, P', \AP, L', R')$ is their product constructed according to \cref{def:product}.
    If $\sem{M}(w)(\ell, \gamma) \neq \bot$, then, by \cref{def:TMDP_semantics}, there exists $p,q \in [0,1]$ such that $\sem{M}(w)(\ell, \gamma) = pq$ and $w, \ell$ is observable with probability $p$ and $R(\ell)(\gamma) = q$.
    If $H$ encodes the reward function of $M$, then it follows from the definition of \prm{} semantics that $\bb{P}_H(\ell, \Gamma^{|\ell|-1}\gamma) = R(\ell)(\gamma)$, where $\Gamma^{|\ell|-1}$ denotes all reward sequences length $|\ell|-1$.
    Notice that $\bb{P}_H(s, \Gamma^n) = \sum_{r \in \Gamma^n} \bb{P}_H(s, r) = 1$ for any $n$ and any $s \in (2^\AP)^*$.
    Suppose now that $\ell = \ell'l$ and $w = w'a$.
    Assuming $H$ is reward-deterministic, there must be states $y,y' \in Y$ such that $\bb{P}_H(\ell, \Gamma^{|\ell|-1}\gamma) = R(\ell)(\gamma) = \tau(y, l, y') = q$ and $\varrho(y, l, y') = \gamma$.
    Moreover, the probability $p$ of observing $w,\ell$ in $M {\times} H$, depends only on the transition probability function $P$ of $M$ whenever the reward probabilities are not in consideration.
    Therefore we have that $\Pr[(w, \ell) \in \mathcal{L}(M)]$ is equivalent to $\Pr[(w', \ell' ) \in \mathcal{L}(M {\times} H)] \cdot P(x, a, x')$ for some $x,x' \in X$.
    Since we have established that $R(\ell)(\gamma) = \tau(y, l, y')$, we may conclude that
    \begin{align*}
        &\sem{M}(w)(\ell,\gamma) = \Pr[(w, \ell) \in \mathcal{L}(M)] \cdot R(\ell)(\gamma) \\
        &= \Pr[(w', \ell') \in \mathcal{L}(M {\times} H)] \cdot P(x, a, x') \cdot \tau(y, l, y') \\
        &= \Pr[(w, \ell) \in \mathcal{L}(M {\times} H)] \\
        &= \sem{M {\times} H}(w, \ell)(\gamma). \qedhere
    \end{align*}
\end{proof}

For an \mdp{} $M$ and a positive integer $n$, define $\mathcal{L}^{\leq n}(M)$ as the set $\bigcup_{k \leq n} \set{(w, \ell) \in A^k {\times} (2^\AP)^k : (w, \ell) \in \mathcal{L}(M)}$.

\begin{lemma}[\cite{TapplerAichernigBacciEichlsederLarsen19}]
\label{lemma:ep_length}
    Two \mdp{}s $M$ and $M'$ with $n$ and at most $n$ states, respectively, are equivalent iff $\sem{M}(w, \ell) = \sem{M'}(w, \ell)$, for all $(w, \ell) \in \mathcal{L}^{\leq n^2 + 1}(M)$.
\end{lemma}

The following result follows from the convergence of $L^\star$ for label-deterministic \mdp{}s, shown in \cite{TapplerAichernigBacciEichlsederLarsen19}.
In that work, the authors show that under uniformly randomized testing strategies, the sampling-based $L^\star$ algorithm converges almost surely in the limit to the \mdp{} under learning.
Our learning algorithm can be seen as a variant of sampling-based $L^\star$ in which sampling is implemented via reinforcement learning.
By resetting the learning rate parameter to very low values at the start of each query and gradually increasing it after a sufficiently long initial period of exploration results in period of essentially random exploration and thus simulates a uniformly random testing strategy.
This allows us to apply the convergence arguments from \cite{TapplerAichernigBacciEichlsederLarsen19} to our setting to assert that our learning procedure converges almost surely in the limit to the \prm{} (or an equivalent \prm{}) encoding the reward function of the \tmdp{} under learning.
In particular, \cref{lemma:ep_length} provides a lower bound on the reinforcement learning episode lengths necessary to achieve convergence in the limit.

\begin{theorem}
\label{thm:convergence}
    Fix a \tmdp{} $M$, and suppose that there exists a \prm{} $H$, with $k$ states, encoding the reward function of $M$.
    Let $H_n$ be the hypothesis \prm{} passed to the $n^{\tn{th}}$ equivalence query, and suppose that the episode length is at least $k^2 + 1$.
    \begin{enumerate}
        \item If every possible $\ell \in (2^\AP)^*$ occurs in a pair $(w, \ell) \in  \mathcal{L}(M)$, then $H_n \to H$ almost surely as $n \to \infty$.
        \item Otherwise, $M {\times} H_n \to M {\times} H$ almost surely as $n \to \infty$.
    \end{enumerate}
\end{theorem}

\Cref{thm:convergence} tells us that \cref{algo:main} almost surely converges to a \prm{} $H$ encoding the reward function of the \tmdp{} $M$ under learning, under the assumption that every possible label sequence is attainable from $M$.
If this assumption does not hold, then we cannot claim that the true \prm{} encoding the reward function has been learned, since we cannot sample the behavior of the system on unobservable traces.
On the other hand, we can say that the learned \prm{} is equivalent to $H$, \emph{modulo the language of $M$}.
Given sufficient interactions, the algorithm will almost surely converge to a \prm{} that is equivalent to $H$ on those label sequences in the language of $M$.
Therefore, optimal strategies for $M {\times} H$ can be learned from the product of $M$ and the learned \prm{}.

The following result justifies the use of \prm{}s by establishing that \prm{}s can provide a more compact representation of systems than might otherwise be modeled by embedding probabilities capturing the stochasticity of a stochastic non-Markovian reward function directly into the environment and using a deterministic reward machine to encapsulate the non-Markovian aspects of the reward function.

\begin{theorem}
\label{thm:deterministic}
    Suppose that $M$ is a \tmdp{} with reward function encoded by a reward-deterministic \prm{} $H$.
    If $M$ has $m$ states and $H$ has $n$ states, then there exists an \tmdp{} $M'$ with $mn$ states with a deterministic reward function encoded by deterministic reward machine $H'$ with $O(2^n)$ states such that the product \mdp{}s $M {\times} H$ and $M' {\times} H'$ are equivalent.
\end{theorem}
\begin{proof}
    Suppose that $M = (X, x_I, A, P, \AP, L, R)$ is a \tmdp{}, $H = (\AP, \Gamma, Y, y_I, \tau, \varrho)$ is a \prm{} encoding $R$.
    We proceed to construct a deterministic \prm{} as follows:
    \begin{enumerate}
        \item We construct an NFA $N$ from $H$ such that rewards are moved into the states and a transition is included in $N$ for every transition of positive probability in $H$.
        \item A DFA $D$ is produced using a subset construction on $N$.
        \item A deterministic reward machine $H'$ is obtained by interchanging deterministic transitions in $D$ with transitions of probability 1 and bringing rewards back out onto the transitions while still keeping them in the states.
    \end{enumerate}
    Note that reward-determinism of $H$ ensures that the determinization step can be carried out without introducing sequences of rewards in $H'$ that were unobtainable in $H$.
     The three steps above can be composed into a single step to obtain a deterministic \prm{} $H' = (\AP, \Gamma, Y', y'_I, \tau', \varrho')$ where
    \begin{itemize}
        \item $Y' = 2^{Y {\times} \Gamma}$ with $y'_I = \set{(y_I, 0)}$,
        \item $\tau'(Y_1, l, Y_2) = 1$ iff $\tau(y, l, y') > 0$ and $\varrho(y, l, y') = \gamma'$, for all $(y,\gamma) \in Y_1$ and $(y', \gamma') \in Y_2$
        \item $\varrho'(Y_1, l, Y_2) = \gamma$ iff $Y_2 \subseteq \set{(y,\gamma) : y \in Y}$.
    \end{itemize}
    Now, we construct, as a modified product of $H$ and $M$, a \tmdp{} $M' = (X', x'_I, A, P', \AP, L', R')$  which simulates the transition probabilities of $H$ and has a deterministic reward function encoded by $H'$:
    \begin{itemize}
        \item $X' = X {\times} Y$ with $x'_I = (x_I, y_I)$,
        \item $P'((x,y), a, (x',y')) = P(x, a, x') \cdot \tau(y, L(x, a, x'), y')$, 
        \item $L'((x,y), a, (x',y')) = L(x, a, x')$,
        \item $R'(\ell)(\gamma) = \bb{P}_{H'}\paren{\ell, \Gamma^{|\ell|-1}\gamma}$.
    \end{itemize}
    Taking the product of $M'$ and $H'$ yields an \mdp{} $M' {\times} H' = (X'', x_I'', A, P'', \AP, L'', R'')$ such that
    \begin{itemize}
        \item $X'' = X' \times Y'$ with $x_I'' = (x'_I, y'_I)$,
        \item $P''((x'_1, y'_1), a, (x'_2,y'_2)) {\ =}$ ${\quad} P'(x'_1, a, x'_2) {\cdot} \tau'(y'_1, L'(x'_1, a, x'_2), y'_2)$,
        \item $L''((x'_1, y'_1), a, (x'_2,y'_2)) = L'(x'_1, a, x'_2)$,
        \item $R''(L''((x'_1, y'_1), a, (x'_2,y'_2))) {=} \varrho'(y'_1, L'(x'_1, a, x'_2), y'_2)$.
    \end{itemize}
    
    What remains to be shown is that the \mdp{}s $M {\times} H$ and $M' {\times} H'$ are equivalent.
    We proceed by induction on $(w, \ell)$.\\
    
    \noindent\emph{Base case: $(w,\ell) = (\varepsilon, \varepsilon)$.}

    It follows immediately from \cref{def:mdp_semantics} that $\sem{M {\times} H}(\varepsilon, \varepsilon)(\gamma) = \sem{M' {\times} H'}(\varepsilon, \varepsilon)(\gamma)$, for any $\gamma \in \Gamma$.\\
    
    \noindent\emph{Inductive case: If $\sem{M {\times} H}(w',\ell') = \sem{M' {\times} H'}(w',\ell')$, for $w' \in A^*$ and $\ell' \in \paren{2^\AP}^*$, then it follows, for all $a \in A$ and $l \in 2^\AP$, that $\sem{M {\times} H}(w'a,\ell'l) = \sem{M' {\times} H'}(w'a,\ell'l)$.}

    Let $w \in A^*$, $\ell \in (2^\AP)^*$, and $\gamma \in \Gamma$, and suppose that $w = w'a$ and $\ell = \ell' l$.
    Then, the following derivation holds for some $x,x' \in X$, $y,y' \in Y$ and $Y_1, Y_2 \subseteq Y$. 
    \begin{align*}
        &\sem{M' {\times} H'}(w, \ell)(\gamma) \\
        &= \pr{(w,\ell) \in \mathcal{L}(M' {\times} H') \tn{ and } \gamma = R''(l)} \\
        &= \pr{(w',\ell') \in \mathcal{L}(M' {\times} H')} \\
        &\quad \cdot P''(((x,y), Y_1), a, ((x',y'), Y_2)) \\
        &= \pr{(w',\ell') \in \mathcal{L}(M' {\times} H')} \cdot P(x, a, x') \\
        &\quad \cdot \tau(y, l, y') \cdot \tau'(Y_1, l, Y_2) \\
        &= \pr{(w',\ell') \in \mathcal{L}(M' {\times} H')} \cdot P(x, a, x') \cdot \tau(y, l, y') \\
        &= \sem{M' {\times} H'}(w', \ell')(\gamma) \cdot P(x, a, x') \cdot \tau(y, l, y') \\
        &= \sem{M {\times} H}(w', \ell')(\gamma) \cdot P(x, a, x') \cdot \tau(y, l, y') \\
        &= \sem{M {\times} H}(w, \ell)(\gamma) \qedhere
    \end{align*}
\end{proof}

\section{Empirical Evaluation}
\label{sec:empirical}
While the key contribution of this paper is theoretical, we implemented the proposed algorithms in Python to evaluate their effectiveness via a number of small-sized benchmarks.
Recall the office gridworld scenario and the corresponding \prm{} shown in  \cref{fig:coffeeworldexample}.
Notice that this \prm{} is neither minimal nor reward-deterministic, yet our algorithm was able to learn the equivalent \prm{}, shown in \cref{fig:learned_office_PRM}, which is both minimal and reward-deterministic.
We experimented with other small-sized environments and \prm{}s, and showed that our algorithm can consistently learn \prm{}s in less than 1 minute on a typical personal laptop. 

\begin{figure}
    \centering
    \begin{tikzpicture}[node distance = 3.5cm]
        \node[state,initial] (0) {$y_0$};
        \node[state] (1) [below of = 0] {$y_1$};
        \node[state] (2) [right of = 0] {$y_2$};
        \node[state] (3) [below of = 2] {$y_3$};
        \path[->] (0) edge [loop above] node {$\_ \mid 1$} node [below] {0} (0);
        \path[->] (0) edge node [above] {$c \mid 1$} node [below] {0} (2);
        \path[->] (0) edge node [left] {$* \mid 1$} node [right] {0} (1);
        \path[->] (2) edge [loop above] node {$\_ \vee c \mid 1$} node [below] {0} (2);
        \path[->] (2) edge [bend right] node [sloped,below] {$o \mid 0.09$} node [sloped,above] {0} (1);
        \path[->] (2) edge [bend left] node [sloped,above] {$* \mid 1$} node [sloped,below] {0} (1);
        \path[->] (3) edge node [sloped, below] {$o \vee \_ \vee * \vee c \mid 1$} node [sloped, above] {0} (1);
        \path[->] (2) edge node [right] {$o \mid 0.91$} node [left] {1} (3);
        \path[->] (1) edge [loop below] node {$o \vee \_ \vee * \vee c \mid 1$} node [above] {0} (1);
    \end{tikzpicture}
    \caption{The \prm{} learned from the office gridworld.}
    \label{fig:learned_office_PRM}
\end{figure}

As expected, the run-time of the algorithm increases exponentially with the number of states in the concept \prm{}; however our experiments showed that the algorithm scales gracefully (linearly) with the number of transitions.
When varying the size of the environment, however, we noticed that certain kinds of environments seemed much more challenging than others.
Upon further inspection, this coincided with the distinction that some of the \prm{}s operate on the predicates over the set of actions while others on the predicates over the set of states (observations). 
The office grid-world example, for instance, is labelled by observations $c$ or $o$, while the actions the agent correspond to the direction of the movement ({\tt n}, {\tt e}, {\tt w}, {\tt s}). 
Of course, the key difference when the \prm{}s operate on predicates over actions is that every label is observable from every state and hence the \mdp{} can generate every word, resulting in efficient membership queries. 
Interestingly, increasing the size of the environment where labels are observable from every state had minimal effect on the performance of the learning.
On the other hand, if increasing the size resulted in very low probabilities of observing certain labels, the algorithm struggled to learn correct \prm{}s.
This difference is a direct consequence of our definition of statistical difference.
The employed notion of difference only distinguishes traces that have some data recorded in the table; it does not differentiate traces if either has no data at all.
Thus, the algorithm cannot tell if the lack of data is because traces have yet to be sampled or if traces are unobservable.
This issue can be alleviated by introducing oracle queries capable of answering questions regarding the observation language of the environment.



\section{Conclusion}
\label{sec:conclusion}
We introduced \emph{probabilistic reward machines} as an extension of reward machines to accommodate non-Markovian and stochastic dynamics. 
We developed a learning algorithm to obtain \prm{}s from underlying decision processes via a combination of model-free reinforcement learning and a probabilistic variant of the $L^\star$ active inference method. 
While this work establishes a theoretical foundation for the use of reward machines to learn non-Markovian stochastic rewards, further research is needed to assess the practicality of our approach. 
We believe that certain tweaks, such as allowing oracle queries on the environmental observation language or using alternative notions of statistical difference and compatibility, will improve performance in practice.

\section*{Acknowledgments}
This research was supported in part by the Air Force Research Laboratory through the Information Directorate’s Information Institute$^\text{®}$ Contract Number FA8750-20-3-1003, FA8750-20-3-1004, SA10032021060389, and SA2020071003-V0237, and the Air Force Office of Scientific Research through Award 20RICOR012, and the National Science Foundation through CAREER Award CCF-1552497 and Award CCF-2106339.


\bibliography{refs}

\end{document}